\pdfoutput=1

\documentclass[11pt]{article}

\usepackage[final]{acl}

\usepackage{times}
\usepackage{latexsym}

\usepackage[most]{tcolorbox}
\usepackage{xcolor}
\usepackage{amssymb}
\usepackage{booktabs}
\usepackage{siunitx}
\usepackage{multirow}
\usepackage{array}
\usepackage{siunitx}

\usepackage[T1]{fontenc}

\usepackage[utf8]{inputenc}

\usepackage{microtype}

\usepackage{inconsolata}

\usepackage{graphicx}
\usepackage{enumitem}

\title{Surprisal reveals diversity gaps in image captioning \\and different scorers change the story}

 \author{Nikolai Ilinykh \and Simon Dobnik \\
         Centre for Linguistic Theory and Studies in Probability (CLASP), \\ Department of Philosophy, Linguistics and Theory of Science (FLoV), \\ University of Gothenburg, Sweden \\         
         \texttt{\{nikolai.ilinykh,simon.dobnik\}@gu.se}
         }

\begin{document}
\maketitle

\begin{abstract}
We quantify linguistic diversity in image captioning with \textbf{surprisal variance} -- the spread of token-level negative log-probabilities within a caption set.
On the MSCOCO test set, we compare five state-of-the-art vision-and-language LLMs, decoded with greedy and nucleus sampling, to human captions.
Measured with a caption-trained n-gram LM, humans display roughly twice the surprisal variance of models, but rescoring the same captions with a general-language model reverses the pattern.
Our analysis introduces the surprisal-based diversity metric for image captioning.
We show that relying on a single scorer can completely invert conclusions, thus, robust diversity evaluation must report surprisal under several scorers.
\end{abstract}

\section{Introduction}

Every person is unique in their linguistic behaviour: the same image may evoke many different yet valid descriptions depending on their different personal contexts, beliefs, attention, and background knowledge.
This ``natural variability'' has been documented in linguistic annotation \citep{plank2014,plank-2022-problem}, inference \citep{pavlick2019}, object naming \citep{silberer2020}, and image captioning \citep{bernardi2016}.
On the other hand, although vision-and-language models may capture fragmets of individual styles, they are normally not trained with capturing diversity in mind. 
They are optimised for maximum-likelihood estimation (MLE), a function that rewards word combinations already dominant in the training corpus and penalises rarer phrasing \citep{devlin2015,dai2017}. 
Decoding tweaks (e.g.\ nucleus sampling \citep{holtzman2020}) or alternative objectives \citep{welleck2020,li-etal-2016-diversity} may mitigate this effect.
Still, transformer‑based models trained with MLE now achieve impressive accuracy on caption benchmarks \citep{alayrac2022,li2023,bai2025qwen25vltechnicalreport,zhu2025}.
However, such benchmarks assume a single ground truth and often focus on targeting this very benchmark, meeting a certain metric score \citep{schlangen-2021-targeting}.
As a result, the diversity of model language is often sidelined -- even though prior studies argue that missing variability is critical for evaluating multi-modal models \citep{castro2016,li2016,vanmiltenburg2018}.
Existing diversity metrics for tasks such as image captioning \citep{shetty2017speakinglanguagematchingmachine,vanmiltenburg2018}, rely on surface counts (length, type–token ratio, distinct‑n) and lack a principled link to linguistic processing.

In this study we examine diversity of image captions through the notion of surprisal, a probability-weighted and context-sensitive probe.
The surprisal, a negative log‑probability of a word in context, has been widely used in linguistic analysis to quantify the processing cost of a message \citep{hale2001,levy2008} and has proved to be a robust metric across languages \citep{pimentel-etal-2021-surprisal,wilcox-etal-2023-testing}.
Generic, more commonly used words and phrases across speakers or models are easier for either to produce, hence any potential differences in the measured surprisal between these two groups naturally reveal any differences in diversity of their linguistic behaviour \citep{gehrmann-etal-2019-gltr,venkatraman-etal-2024-gpt,xu-etal-2024-detecting}.
We extend this line of work from text‑only settings to image captions. 

We estimate the probabilities used to measure surprisal using two 
probabilistic scorers (i) an n-gram model trained on a large corpus consisting of a balanced collection of human and model generated image captions (in-domain expert), and (ii) GPT-2 \citep{radford2019language}, a general-purpose transformer trained on broad English.
As we do not have access to the true population of all possible captions for images, we follow earlier work \citep{SMITH2013302,wilcox-etal-2023-testing,giulianelli-etal-2023-comes} and approximate the underlying distribution with two different language models.
%
%
Inspired by the scorer-sensitivity findings of \citet{arora-etal-2022-estimating}, our dual‑scorer design tests whether conclusions about diversity change with the change of the evaluator of surprisal.

Our research questions are:
\begin{enumerate}[itemsep=0.1pt,topsep=10pt]
    \item Are human descriptions of individual images more diverse than those generated by language models in terms of the estimated surprisal of the in-domain unbiased scorer? 
    \item Do predictions of this scorer align to those of a general language model scorer which has been trained on open text?
\end{enumerate}

As artificial describers (description generators) we analyse five state-of-the-art vision-and-language models and compare their descriptions with human captions.
Like \citet{zamaraeva2025comparingllmgeneratedhumanauthorednews}, who used linguistic theory to identify syntactic gaps in LLM-generated news compared to human-authored news, we resort to the linguistic theory of surprisal to evaluate multi‑modal language, showing how scorer choice interacts with caption distributions.
Our study reveals a clear difference in surprisal and therefore diversity between human and model captions, and, importantly, that the direction of this diversity flips with the chosen scorer.
These differences between 
evaluation scorers underscore the need for future benchmarks to include surprisal-based diversity scores computed under several, well‑motivated scorers.

\section{Materials and methods}

\subsection{Data}

We use the Karpathy test-split of MSCOCO \citep{lin2014}, which consists of $5000$ images each paired with five independent human captions ($25000$ descriptions total).
MSCOCO is one of the popular and commonly used image captioning benchmarks. It provides multiple human references per image -- exactly what we need to quantify how diversely an image can be described by humans and models\footnote{We note that MSCOCO is widely incorporated in training of multi-modal LLMs.}.

\subsection{Models}

We use five vision–language models to generate one description per image: Qwen2.5‑VL‑72B‑Instruct \citep{bai2025qwen25vltechnicalreport}, InternVL3-78B‑78B‑Instruct \citep{zhu2025}, Llama‑4‑Scout‑17B‑16E‑Instruct\footnote{\url{https://ai.meta.com/blog/llama-4-multimodal-intelligence/}}, Claude Sonnet 4\footnote{\url{https://www.anthropic.com/news/claude-4}}, and GPT‑4o \citep{openai2024gpt4ocard}.
These include both closed‑source systems and large open‑source models, all ranking among the top performers on the MMMU benchmark \citep{yue2024mmmumassivemultidisciplinemultimodal}.
Our choice of five models is intended to correspond to five human describers per image in the MS COCO dataset \citep{lin2014}. 
This way we attempt to mirror a scenario of difference in experience between between different human describers, as the models 
are based on different architectures and training regimes.
More on technical details can be found in Appendix A.
%
%
We produce two different sets of texts by running models with two decoding algorithms: greedy search and nucleus sampling \citep{holtzman2020}.
Different decoding strategies let us assess model captions in a more nuanced way since decoding has a strong effect on the output in NLG \citep{info12090355}.

\subsection{Scorers of caption probability distributions}

\paragraph{Overview}
For each image in our dataset, we train a single Kneser–Ney‑smoothed n‑gram model (bi- and tri-) on the \emph{union} of all human and model captions \emph{except} those associated with that target image which will be our test set. 
As the probabilities estimated by this model are based on both human and artificially generated captions we use this ``balanced'' language model to score each of the the held‑out captions (five human, five machine), giving us one surprisal value per caption.
Our zero hypothesis is that this model by virtue of being balanced) will predict no differences in surprisal value between human and model generated captions of individual images. 

\paragraph{Training details}
We train both bigram and trigram models to check that our findings are robust to context size.
All models are implemented with NLTK’s \texttt{KneserNeyInterpolated} language model API \citep{bird2009,kneserney}.
With 5000 images $\times$ 10 captions each, each leave‑one‑image‑out model is trained on \((5000-1)\times10 = 49\,990\) captions.
The vocabulary of the model is built from the training pool (excluding the target image).
%
%
Pooling human and machine captions ensures both groups contribute the same n‑gram types and counts, so rare‑word penalties are applied equally.
Also, since all captions are scored by the same language model, per‑token surprisal is directly comparable between human and model outputs.
%
Even if models saw MSCOCO during pre‑training, our leave‑one‑image‑out n‑gram setup excludes each target image’s captions from training, so repeated references are still treated as new, ensuring surprisal reflects style rather than memorisation.

\subsection{Surprisal as evaluation metric}
In psycholinguistics and computational linguistics, \textit{surprisal} \citep{shannon} is a well‑established measure of the information conveyed by a word in context.
Formally, the surprisal of a linguistic unit $w_t$ given preceding units $w_{<t}$ is defined as the negative log‑likelihood of the word conditioned on previous context \citep{hale2001,levy2008}:

\begin{align}
    I(w_t) = - \log P_{\theta}(w_t \mid w_{<t}),
\end{align}

\noindent where $P$ is the underlying probability distribution.
Intuitively, treating units as words, predictable words carry less information (lower surprisal), while unexpected words convey more (higher surprisal).
We use surprisal to quantify linguistic unpredictability in image captions under a model $\theta$ trained to approximate $P$.
We compute word‑level surprisals and average them across each caption to estimate how predictable the caption is under a given evaluator.
Captions with higher average surprisal are considered more lexically or structurally novel, while captions with lower surprisal follow patterns captured in $P_\theta$.

\section{Experiments and results}

Before moving to our surprisal-based experiments, we first ask if model captions even look like human captions on lexical level.
We compute several metrics described in \citet{vanmiltenburg2018}; the results are shown in Table~\ref{tab:overall-metrics}.
\begin{table}[ht]
\small
\centering
\setlength{\tabcolsep}{5pt}
\begin{tabular}{
>{\centering\arraybackslash}p{1cm}
>{\centering\arraybackslash}p{2cm}
>{\centering\arraybackslash}p{1cm}
>{\centering\arraybackslash}p{1cm}
>{\centering\arraybackslash}p{1cm}
}
\toprule
Source & ASL $\pm$ SDSL & \#Types & TTR1 & TTR2 \\
\midrule
Human          & $10.44 \pm 2.36$  & 7,252  & 0.28 & 0.66 \\
\texttt{greedy}   & $39.19 \pm 53.35$ & 10,783 & 0.29 & 0.66 \\
\texttt{nucleus}  & $40.00 \pm 27.94$ & 13,082 & 0.33 & 0.73 \\
\bottomrule
\end{tabular}
\caption{Overall lexical statistics for human captions and all model captions combined under \texttt{greedy} and \texttt{nucleus} decoding. ASL = Average Sentence Length (tokens per caption), SDSL = Standard Deviation of Sentence Length, \#Types = number of unique word types, TTR1 = type-token ratio (per 1000 tokens), TTR2 = bigram type–token ratio (per 1000 bigrams).}
\label{tab:overall-metrics}
\end{table}
\begin{table*}[ht!]
\centering
\setlength{\tabcolsep}{5pt}

\begin{tabular}{
>{\centering\arraybackslash}p{2.7cm}    
>{\centering\arraybackslash}p{2.0cm}    
S[table-format=2.3]@{$\,\pm\,$}S[table-format=2.3]   
S[table-format=2.3]@{$\,\pm\,$}S[table-format=2.3]   
>{\centering\arraybackslash}p{1.5cm}      
>{\centering\arraybackslash}p{1cm}                                 
}
\toprule
\multirow{2}{*}{\centering $P_\theta$} &
\multirow{2}{*}{\centering Data} &
\multicolumn{2}{c}{Human} &
\multicolumn{2}{c}{Model} &
\multirow{2}{*}{$t$-value} &
\multirow{2}{*}{$d_z$} \\
\cmidrule(lr){3-4}\cmidrule(lr){5-6}
 & & \multicolumn{2}{c}{Mean $\pm$ SD} &
     \multicolumn{2}{c}{Mean $\pm$ SD} & & \\
\midrule
Bi‑gram LM &
$\mathcal{H} \cup \mathcal{M}_{\text{greedy}}$ &
5.20 & 5.04 & 2.17 & 2.00 & {40.88\textsuperscript{***}} & 0.58 \\

Tri‑gram LM &
$\mathcal{H} \cup \mathcal{M}_{\text{greedy}}$ &
7.39 & 6.04 & 3.71 & 3.04 & {39.45\textsuperscript{***}} & 0.56 \\

Bi‑gram LM &
$\mathcal{H} \cup \mathcal{M}_{\text{nucleus}}$ &
5.03 & 4.95 & 4.08 & 3.30 & {11.50\textsuperscript{***}} & 0.16 \\

Tri‑gram LM &
$\mathcal{H} \cup \mathcal{M}_{\text{nucleus}}$ &
7.34 & 6.09 & 6.74 & 4.49 & {5.70\textsuperscript{***}} & 0.08 \\

\midrule

GPT‑2 (small) &
$\mathcal{H} \cup \mathcal{M}_{\text{greedy}}$ &
1.565 & 1.521 & 2.015 & 1.323 & {-16.37\textsuperscript{***}} & 0.23 \\

GPT‑2 (medium) &
$\mathcal{H} \cup \mathcal{M}_{\text{greedy}}$ &
1.485 & 1.420 & 2.016 & 1.316 & {-20.31\textsuperscript{***}} & 0.29 \\

GPT‑2 (XL) &
$\mathcal{H} \cup \mathcal{M}_{\text{greedy}}$ &
1.438 & 1.344 & 1.954 & 1.243 & {-20.85\textsuperscript{***}} & 0.29 \\

GPT‑2 (small) &
$\mathcal{H} \cup \mathcal{M}_{\text{nucleus}}$ &
1.565 & 1.521 & 2.906 & 24.788 & {-3.82\textsuperscript{***}} & 0.05 \\

GPT‑2 (medium) &
$\mathcal{H} \cup \mathcal{M}_{\text{nucleus}}$ &
1.485 & 1.420 & 2.903 & 24.824 & {-4.03\textsuperscript{***}} & 0.06 \\

GPT‑2 (XL) &
$\mathcal{H} \cup \mathcal{M}_{\text{nucleus}}$ &
1.438 & 1.344 & 2.911 & 23.710 & {-4.39\textsuperscript{***}} & 0.06 \\

\bottomrule
\end{tabular}

\caption{Variance in surprisal for human and model image descriptions
($n = 5{,}000$ images; paired $t$‑test, $df = 4\,999$).
\textsuperscript{***}\,indicates $p < .001$.
$d_z \ge 0.50$ indicates moderate effect size, $d_z < 0.20$ indicates small effect size.
For n-gram models, training data for each language‑model order $n$ (under ``Data'' column) is the union of the human
corpus $\mathcal{H}$ and machine‑generated texts $\mathcal{M}$ produced
with the indicated decoding strategy (greedy or nucleus sampling).
The same data was used to test pre-trained GPT-2.}
\label{tab:res1}
\end{table*}
A table with per model lexical diversity metrics is available in Appendix C. 
The results show that human and model captions look different.
Humans produce concise, about $10$ words long captions with little variation, whereas models spin out strings three to four times longer and far less consistent in length.
Despite this gap, models list far more unique words than humans, possibly because their captions are several times longer.
However, once normalised by output length (TTR metrics), the groups converge.
This interesting result suggests that the apparent richness of model vocabulary is largely an artifact of their verbosity.
When captions are several times longer, the chance of introducing new word types naturally increases, even if much of the added material is repetitive.
Once we control for output length, however, the underlying lexical behaviour looks remarkably similar.
This convergence implies that models, like humans, operate within a comparable core vocabulary for the task, but their training encourages them to elaborate rather than compress.
In other words, verbosity does not necessarily translate to greater genuine lexical creativity -- instead, it may reflect a tendency to ``pad'' responses with familiar constructions, a behaviour aligned with next-word prediction training objectives rather than pragmatic efficiency in describing images.

Overall, the results in Table~\ref{tab:res1} demonstrate that models lengthen their captions mostly by repeating words known to them, so the proportion of genuinely new vocabulary remains virtually the same as in the brief human captions.
Such extensive text generation is not pragmatically required as demonstrated by the much shorter human-generated texts for the captioning task.
These findings can also be linked to differences in captioning guidelines: those presented to humans for MSCOCO data collection (i.e., ``focus on visual content'') versus those followed by instruction-tuned models, which were trained to pursue a different goal from that of humans.
The reason for the models’ verbosity could thus be their tendency to generate texts features different from those found in human-generated texts, including syntactic structures and vocabulary diversity \citep{Mu_oz_Ortiz_2024} as well as grammatical differences \citep{zamaraeva2025comparingllmgeneratedhumanauthorednews}.
The effect of the task provided to the image describers on the properties and features of the produced captions is also important \citep{ilinykh-etal-2018-task}.
\begin{figure}[t]
\centering
\scriptsize 
\begin{minipage}[t]{0.2\textwidth}
    \centering
    \includegraphics[width=\linewidth]{coco-example.jpg}

\end{minipage}\hfill
\begin{minipage}[t]{0.4\textwidth}
    \textbf{Human caption:}\\
    ``The two monitors are sitting among the laptops.''\\[0.8em]
    \textbf{GPT4o:}\\
    ``The workspace setup includes two laptops, two external monitors, two full-size keyboards, and various accessories. One laptop on the left is open and displaying a green-themed image, while another on the right sits elevated on a stand showing a white car. The two central monitors display websites and applications, likely for multitasking. The desk also holds a pair of headphones, a computer mouse for each setup, a smartphone, and other small electronics, indicating a high-performance workstation for media, development, or design tasks.''
\end{minipage}
\caption{An example image from MSCOCO test set with one human reference and one model caption.}
\label{fig:workstation}
\end{figure}
Two captions can be similar or different in terms of length and number of types, yet be radically different in how predictable or information‑rich they are.
Look at the example in Figure~\ref{fig:workstation}.
Both captions mention the same objects (monitors, laptops) and share word types, but the model’s version explores far less frequent phrases (``workspace setup''. ``green‑themed image'') and long chains of modifiers that could carry a different amount of information content.
Surprisal under scorers can capture these probability‑weighted differences that surface counts alone cannot.
In other words, surprisal lets us measure not just how many different words appear, but how unexpectedly they are combined.

One striking difference is that models tend to \textit{overdescribe} visual information exhaustively, while humans focus on particular elements.
This selectivity may also allow for greater diversity since different humans can choose to highlight different aspects of an image.
If humans are behaving pragmatically, they generate optimal (concise) captions, and anything longer would be pragmatically inappropriate.
Multi-modal LLMs, by contrast, struggle to capture captioning intent because they produce much longer texts.
This is an interesting finding as it raises the question of what constitutes a good model-generated description.
We leave this discussion for future work.
Next, we move to our main experiments.

\subsection{Surprisal within groups: in-domain}
\label{subsec:ngramsurprisal}

We compute surprisal variance across five captions per group for each image and use a paired t‑test \citep{student08ttest} to check if human and model variances differ.
As shown in Table~\ref{tab:res1}, for both n-gram orders under both decoding methods, human descriptions have shown roughly two times higher variance in surprisal than model descriptions.
Switching to \texttt{nucleus} narrows the gap, but only slightly as humans still remain the more variable group.
This result suggests that different models (although the differ in size, training data and performance on benchmarks) tend to generate image descriptions with similar information content, whereas humans offer more variance in the information they express.
Per model descriptive analysis is available in Appendix B.


\subsection{Surprisal within groups: general-domain}

The analysis in Section~\ref{subsec:ngramsurprisal} characterises surprisal variance under the probability distribution $P_\theta$ that
is estimated by a simple n-gram language model trained in-domain.
While these models provide interpretable baselines, their probability estimates are limited by local context and the sparsity of the caption corpus.
To test whether our findings generalise under a richer representation of linguistic structure, we replace the caption‑trained n-gram models with a large pre‑trained language model (GPT-2 \citep{radford2019language}).
This substitution changes the underlying probability distribution to the one trained on a much broader and more expressive corpus, that is more of a general English corpus rather than caption-related one.
By doing so, we can examine whether the observed human–model differences in surprisal variance persist when surprisal is computed under a generally more powerful, but different model of language.
To compute surprisal with GPT-2, we use codebase provided by \citet{oh-etal-2024-frequency}\footnote{The codebase is available here: \url{https://github.com/byungdoh/llm_surprisal}}.
Recently, previous research has used LLMs as surprisal scorers in the context of second-language writing development \citep{hu-cong-2025-modeling}.
According to Table~\ref{tab:res1}, when surprisal is estimated with the general English language scorer, the trends reverse.
Across all GPT‑2 configurations, human surprisal variance was lower than model variance, with significant paired differences.
One possible explanation for this reversal lies in the training objectives and data distributions of the respective models.
The in-domain n-gram model trained only on captions is highly sensitive to local lexical patterns and reflects the narrow regularities of the captioning domain.
In contrast, GPT-2 is trained on broad and heterogeneous corpora with the objective of next-word prediction.
Under this objective, model-generated captions may appear more variable because they deviate from the stylistic and structural norms GPT-2 has internalised from general English texts, whereas human captions -- short, formulaic, and pragmatically efficient -- are closer to those norms and thus exhibit lower surprisal variance.
In this sense, the two scorers are complementary rather than directly comparable: the n-gram model emphasises within-domain variation, while GPT-2 highlights divergence from general-purpose English usage.
The methodological implication is that conclusions about human–model variability depend strongly on the choice of reference probability distribution.

\section{Conclusion}

Our study provides three main messages.
First, using surprisal as a measure of diversity gap in image captioning requires reporting under multiple scorers.
Doing so will (i) prevent over-interpretation of scores tied to one distribution, and (ii) encourage future models to match human-level variability inside the caption genre rather than describing images in general language.
The apparent reversal in variance of surprisal within describer groups is a diagnostic of whose expectations one chooses as their scorer.
The two different scorers we employ in this short study should be seen as two complementary metrics.
Our work is thus helpful when deciding which scorer to choose given a particular task.


\section*{Limitations}

Our analysis centers around two probability models to compute surprisal: the caption-domain n‑gram language model and GPT‑2.
Using additional scorers, including larger or instruction‑tuned LLMs will test whether the observed patterns generalise under different scorers.
We use MSCOCO Karpathy test split because it offers five independent human references per image.
Running our experiments on other datasets with different stylistic conventions and multiple references such as Flickr 30k \citep{plummer2016flickr30kentitiescollectingregiontophrase} would test robustness of our method.
Linking variance in surprisal to accuracy metrics such as BERTScore \citep{zhang2020} or MoverScore \citep{zhao2019} will show whether diversity gains align with or trade off against semantic quality.
Finally, both captions and scorers are English‑based; the interaction between genre‑specific and general‑language scorers may differ in morphologically richer or typologically distant languages.

\section*{Acknowledgments}
The work reported in this paper has been supported by a grant from the Swedish Research Council (VR project 2014-39) for the establishment of the Centre for Linguistic Theory and Studies in Probability (CLASP) at the University of Gothenburg.
The computations and data storage were enabled by resources provided by the National Academic Infrastructure for Supercomputing in Sweden (NAISS), partially funded by the Swedish Research Council through grant agreement no. 2022-06725.

\bibliography{references}

\appendix

\section{Technical details}
\label{sec:appendixA}


Each model was prompted with the following instructions:

\begin{tcolorbox}[colback=gray!10!white, colframe=gray!50!black, 
                  title=Prompt Setup, fonttitle=\bfseries]
\textbf{System prompt:}\\
You are a helpful annotator tasked with describing images. You will see one image and you will be asked to describe it.

\vspace{1em}
\textbf{User prompt:}\\
Describe all the important parts of the scene. Do not start the sentence with ``There is''. Do not describe unimportant details. Do not describe things that might have happened in the future or past. Do not describe what a person might say. Do not give people proper names. The sentence should contain at least 8 words.
\end{tcolorbox}

These instructions replicate those given to human annotators in the MS COCO captioning dataset \citep{chen2015microsoftcococaptionsdata}.
Each model received both the prompt and the image as input.

The open‑source models were loaded from HuggingFace repositories.
We used \texttt{OpenGVLab/InternVL3-78B-78B-Instruct} and \texttt{meta-llama/Llama-4-Scout-17B-16E-Instruct} and served them locally using \texttt{vLLM}~\citep{kwon2023efficient} for efficient inference.
We used the official repository of Qwen2.5‑VL\footnote{\url{https://github.com/QwenLM/Qwen2.5-VL.git}} to run \texttt{Qwen/Qwen2.5-VL-72B-Instruct}.
We used \texttt{claude-sonnet-4-20250514} accessed through the Anthropic API\footnote{\url{https://docs.anthropic.com/en/api/messages}}.
GPT‑4o was accessed through the OpenAI API\footnote{\url{https://platform.openai.com/docs/api-reference/}}, using version \texttt{gpt-4o-2024-08-06}.

All open‑source models were run offline on four NVIDIA A100 GPUs ($80$ GB each, except Qwen2.5-VL which used $4$ $\times$ $40$ GB).
InternVL3-78B and Llama-4-Scout were run with a maximum context length of $8192$ tokens.
Qwen2.5-VL was run with a maximum output length of $300$ tokens and a minimum of $50$ new tokens.
A full pass over $5000$ images took approximately $12$ -- $14$ hours per model.
To ensure comparability across models (including those without beam search support), we used two decoding modes for all models:

\begin{itemize}
    \item \texttt{greedy search}:\\ temperature $0.0$, top\_p $0.0$
    \item \texttt{nucleus sampling}:\\ temperature $1.0$, top\_p $0.92$
\end{itemize}

Each generation was capped at $300$ tokens. All runs used a fixed random seed ($42$) to ensure reproducibility.

\paragraph{Pre-processing}
We observed that Qwen2.5‑VL occasionally hallucinates, producing emoji unicodes, Chinese characters, extra spaces, and line breaks.
To clean these artifacts, we process all models' outputs so that they contain only ASCII letters, digits, standard punctuation, and spaces.
We tokenise all model‑generated captions with PTBTokenizer (using the original MS COCO evaluation code\footnote{\url{[https://github.com/tylin/coco-caption}}) so that they share exactly the same tokenisation as the human references.

\section{Variance in surprisal per model}
\label{sec:appendixB}

\begin{table*}[ht]
\centering
\setlength{\tabcolsep}{4pt}

\begin{tabular}{
>{\centering\arraybackslash}p{3.7cm}   
>{\centering\arraybackslash}p{4cm}   
S[table-format=2.3]                   
S[table-format=1.3]@{$\,\pm\,$}S[table-format=1.3]  
}
\toprule
Configuration & Model &
{Mean surprisal} &
\multicolumn{2}{c}{Variance $\pm$ SD} \\
\midrule
Bi-gram LM, {\texttt{greedy}} & Claude Sonnet 4     & 8.864 & 3.575 & 1.891 \\
Bi-gram LM, {\texttt{greedy}} & GPT4o       & 8.732 & 2.551 & 1.597 \\
Bi-gram LM, {\texttt{greedy}} & InternVL3-78B   & 7.995 & 3.130 & 1.769 \\
Bi-gram LM, {\texttt{greedy}} & Llama 4 Scout-17B-16E & 8.175 & 1.442 & 1.201 \\
Bi-gram LM, {\texttt{greedy}} & Qwen2.5-VL-72B & 10.159& 3.124 & 1.767 \\
\addlinespace
Tri-gram LM, {\texttt{greedy}} & Claude Sonnet 4     & 8.385 & 6.106 & 2.471 \\
Tri-gram LM, {\texttt{greedy}} & GPT4o       & 8.179 & 4.460 & 2.112 \\
Tri-gram LM, {\texttt{greedy}} & InternVL3-78B   & 7.042 & 5.444 & 2.333 \\
Tri-gram LM, {\texttt{greedy}} & Llama 4 Scout-17B-16E & 7.316 & 2.876 & 1.696 \\
Tri-gram LM, {\texttt{greedy}} & Qwen2.5-VL-72B & 9.964 & 5.047 & 2.246 \\
\addlinespace
Bi-gram LM, {\texttt{nucleus}} & Claude Sonnet 4    & 8.742 & 3.311 & 1.820 \\
Bi-gram LM, {\texttt{nucleus}} & GPT4o      & 9.146 & 2.742 & 1.656 \\
Bi-gram LM, {\texttt{nucleus}} & InternVL3-78B  & 8.837 & 3.866 & 1.966 \\
Bi-gram LM, {\texttt{nucleus}} & Llama 4 Scout-17B-16E& 8.458 & 1.310 & 1.145 \\
Bi-gram LM, {\texttt{nucleus}} & Qwen2.5-VL-72B& 12.086& 4.219 & 2.054 \\
\addlinespace
Tri-gram LM, {\texttt{nucleus}} & Claude Sonnet 4    & 8.269 & 5.763 & 2.401 \\
Tri-gram LM, {\texttt{nucleus}} & GPT4o      & 8.817 & 4.695 & 2.167 \\
Tri-gram LM, {\texttt{nucleus}} & InternVL3-78B  & 8.329 & 6.253 & 2.501 \\
Tri-gram LM, {\texttt{nucleus}} & Llama 4 Scout-17B-16E& 7.840 & 2.609 & 1.615 \\
Tri-gram LM, {\texttt{nucleus}} & Qwen2.5-VL-72B& 12.667& 5.197 & 2.280 \\
\bottomrule
\end{tabular}

\caption{Per‑model surprisal statistics across images. For each configuration and model, the table reports the mean surprisal and the variance in surprisal across images with its standard deviation (shown as variance ± SD), computed over 5,000 images.}
\label{tab:res-models}
\end{table*}

Descriptively, the models show clear differences in how predictable or variable their outputs are.
Across all decoding configurations, Llama 4 Scout-17B-16E shows the lowest variance in surprisal suggesting more uniform descriptions across images.
In contrast, Qwen2.5-VL-72B shows the highest mean surprisal and higher variance values.
Claude Sonnet 4 and GPT4o fall in between but show an increase in variance from bigram to trigram contexts.

\section{Lexical-based diversity metrics per model}
\label{sec:appendixC}

\begin{table*}[ht]
\centering
\setlength{\tabcolsep}{5pt}
\begin{tabular}{
>{\centering\arraybackslash}p{4cm}
>{\centering\arraybackslash}p{1.6cm}
>{\centering\arraybackslash}p{1.6cm}
>{\centering\arraybackslash}p{2cm}
>{\centering\arraybackslash}p{1.4cm}
>{\centering\arraybackslash}p{1.4cm}
}
\toprule
Model & ASL & SDSL & \#Types & TTR1 & TTR2 \\
\midrule
\multicolumn{6}{c}{\texttt{greedy}} \\
\midrule
Claude Sonnet 4     & 24.41 & 7.09  & 5,431 & 0.40 & 0.79 \\
GPT4o       & 35.59 & 15.63 & 6,507 & 0.38 & 0.79 \\
InternVL3-78B   & 15.64 & 7.48  & 3,960 & 0.38 & 0.74 \\
Llama 4 Scout-17B-16E & 77.08 & 107.82 & 6,702 & 0.27 & 0.64 \\
Qwen2.5-VL-72B & 43.23 & 3.99  & 7,021 & 0.46 & 0.86 \\
\midrule
\multicolumn{6}{c}{\texttt{nucleus}} \\
\midrule
Claude Sonnet 4     & 24.64 & 7.30  & 5,460 & 0.40 & 0.78 \\
GPT4o       & 35.39 & 15.70 & 7,031 & 0.40 & 0.81 \\
InternVL3-78B   & 17.77 & 10.36 & 4,991 & 0.41 & 0.80 \\
Llama 4 Scout-17B-16E & 79.04 & 33.79 & 7,570 & 0.32 & 0.73 \\
Qwen2.5-VL-72B & 43.15 & 8.32  & 8,910 & 0.51 & 0.92 \\
\bottomrule
\end{tabular}
\caption{Per‑model lexical statistics under greedy and nucleus decoding. ASL = Average Sentence Length (tokens per caption), SDSL = Standard Deviation of Sentence Length, \#Types = number of unique word types, TTR1 = type–token ratio (per 1000 tokens), TTR2 = bigram type–token ratio (per 1000 bigrams).}
\label{tab:permodel-metrics}
\end{table*}


\end{document}